%% file: main.tex
\documentclass[10pt,twocolumn,letterpaper]{article}

\usepackage{cvpr}              
\input{preamble}
\definecolor{cvprblue}{rgb}{0.21,0.49,0.74}
\usepackage[pagebackref,breaklinks,colorlinks,allcolors=cvprblue]{hyperref}
\usepackage[utf8]{inputenc} 
\usepackage[T1]{fontenc}    
\usepackage{hyperref}       
\usepackage{url}            
\usepackage{booktabs}       
\usepackage{amsfonts}       
\usepackage{nicefrac}       
\usepackage{microtype}      
\usepackage{xcolor}         
\usepackage{acronym}
\usepackage{graphicx}
\usepackage{wrapfig}

\usepackage{amsmath}
\usepackage{amssymb}
\usepackage{mathtools}
\usepackage{amsthm}
\usepackage{algorithm}
\usepackage{algpseudocode}
\usepackage{setspace} 
\usepackage{enumitem}
\usepackage{acronym}
\usepackage{xspace}

\usepackage{graphicx}    
\usepackage{array}       
\usepackage[misc]{ifsym} 
\usepackage{makecell}    
\usepackage{booktabs}    
\usepackage{setspace}    

\usepackage{booktabs}
\usepackage{siunitx}
\usepackage{multirow}
\usepackage{indentfirst}
\usepackage{subcaption}
\usepackage[table]{xcolor}
\usepackage{pifont}
\usepackage{lineno}
\def\paperID{} 
\def\confName{CVPR}
\def\confYear{2026}

\title{DriveLaW: Unifying Planning and Video Generation in a Latent Driving World}

\author{
Tianze Xia$^{1,2*}$	 \quad
Yongkang Li$^{1,2*}$ \quad
Lijun Zhou$^{2*}$ \quad
Jingfeng Yao$^{1}$ \quad
Kaixin Xiong$^{2}$ \quad
Haiyang Sun$^{2\dagger}$	\quad\\
Bing Wang$^{2}$ \quad
Kun Ma$^{2}$ \quad
Guang Chen$^{2}$ \quad
Hangjun Ye$^{2}$ \quad
Wenyu Liu$^{1}$ \quad
Xinggang Wang$^{1}$\textsuperscript{\Letter} \\
\\
$^{1}$Huazhong University of Science and Technology \quad $^{2}$Xiaomi EV\\
}

\begin{document}
\maketitle

\let\thefootnote\relax\footnotetext{
$^*$~Equal contribution: T. Xia (\url{xiatianze@hust.edu.cn}), Y. Li (\url{liyk@hust.edu.cn}) and L. Zhou (\url{zhoulijun16@mails.ucas.edu.cn}). \quad
$^\dagger$~Project lead: H. Sun (\url{AmazingRoad@163.com}). \quad
\textsuperscript{\Letter}~Corresponding author: X. Wang (\url{xgwang@hust.edu.cn}).
}

\input{sec/0_abstract}    
\input{sec/1_intro}

\input{sec/2_related_works}
\input{sec/3_method}

\input{sec/4_experiment}
\input{sec/5_conclusion}

{
    \small
    \bibliographystyle{ieeenat_fullname}
    \bibliography{main}
}

\input{sec/X_suppl}

\end{document}

%% file: sec/0_abstract.tex
\begin{abstract}
World models have become crucial for autonomous driving, as they learn how scenarios evolve over time to address the long-tail challenges of the real world. However, current approaches relegate world models to limited roles: they operate within ostensibly unified architectures that still keep world prediction and motion planning as decoupled processes. To bridge this gap, we propose \textbf{DriveLaW}, a novel paradigm that unifies video generation and motion planning. By directly injecting the latent representation from its video generator into the planner, DriveLaW ensures inherent consistency between high-fidelity future generation and reliable trajectory planning. Specifically, DriveLaW consists of two core components: DriveLaW-Video, our powerful world model that generates high-fidelity forecasting with expressive latent representations, and DriveLaW-Act, a diffusion planner that generates consistent and reliable trajectories from the latent of DriveLaW-Video, with both components optimized by a three-stage progressive training strategy. New state-of-the-art results across both tasks demonstrate the power of our unified paradigm. DriveLaW not only significantly advances video prediction, surpassing the previous best-performing work by 33.3\% in FID and 1.8\% in FVD, but also sets a new record on the NAVSIM planning benchmark. Code available at \url{https://github.com/xiaomi-research/drivelaw}.



\end{abstract}


%% file: sec/1_intro.tex
\section{Introduction}
\label{sec:intro}

Autonomous driving has advanced rapidly in recent years, driven by significant progress in perception~\cite{li2024bevformer, yang2023bevformer, liao2022maptr, huang2021bevdet} and planning~\cite{hu2023planning, jiang2023vad, liao2025diffusiondrive, li2025recogdrive}. However, existing systems remain brittle in long-tail and rare scenarios, degrading closed-loop performance. Recent work has introduced world models that forecast the future evolution of driving scenes from past multi-view observations and ego-state to address long-tail brittleness. They can synthesize downstream task data~\cite{gao2024vista,hu2023gaia,gao2023magicdrive,wang2024drivedreamer,li2024drivingdiffusion} for rare scenarios, enabling policy learning in simulation~\cite{gao2025rad,yang2025resim,li2025omninwm}, and provide auxiliary future supervision~\cite{li2025drivevla,li2024enhancing,zhao2025forecasting,min2024driveworld,wang2024driving}, both of which contribute to improving generalization and robustness under distribution shift.



Despite the generalization gains from learning physical regularities through large-scale video generation, current world models’ contributions to autonomous driving planning remain either indirect or merely parallel to the planner, rather than tightly coupled with decision making. Specifically, in terms of their role in planning, the existing world-model approaches fall into three categories. First, \emph{world-model simulators}~\cite{li2025uniscene,guo2025genesis,zhou2024hugsim,yang2025resim,gao2024vista,hu2023gaia} synthesize downstream data or serve as closed-loop environments to guide policy learning, which is indirect and does not transmit the model’s physical understanding into the planner’s state. Second, \emph{world-model supervision}~\cite{li2025drivevla,li2024enhancing,zheng2024occworld,wang2024occsora} predicts future visual or affordance signals to supervise future frames, occupancy, or trajectories, improving foresight but keeping planning externally specified. Third,  \emph{unified world-model}~\cite{zhang2025epona,bartoccioni2025vavim,zeng2025futuresightdrive} co-generates videos and trajectories, yielding tighter coupling between perception and control and improved temporal consistency. However, current instantiations of this paradigm still fall short of fully realizing the intended tight coupling. Methods such as Epona~\cite{zhang2025epona} and DriveVLA-W0~\cite{li2025drivevla} decouple generation and planning, training the video generator and the policy head as separate modules. Despite evidence that video generators encode strong world understanding and perceptual priors, these approaches do not leverage the generator’s internal latents as the planning state, leaving a gap between visual imagination and action selection.

In this paper, we propose DriveLaW, a latent world model that unifies generation and planning through a shared latent-space representation. Instead of treating generation and planning as two parallel modules, we chain generation and planning inspired by Genie envisioner~\cite{liao2025genie} as shown in Fig.~\ref{fig:framework}, leveraging the strong driving representations learned by the video generator from large-scale video generation to perform trajectory planning. Latent features from the video generator, learned on large-scale driving videos, encapsulate scene semantics, agent dynamics, and physical regularities in a compact representation. The Action DiT (Diffusion Transformer~\cite{peebles2023scalable}), conditioned on these latents, generates temporally distribution-robust trajectories and improves closed-loop stability. In comparison, our chained design offers three advantages over parallel baselines: (1) it fully exploits representations learned from large-scale video pretraining, unifying generation and planning in a shared latent space; (2) its training regime avoids gradient interference between the video generator and the planner; (3) cascading ensures consistency between generated visual detail and the planned trajectories. However, high-fidelity video synthesis and real-time stable planning are inherently in tension. To address this, we design \textit{DriveLaW-Video} with a spatiotemporal VAE and an efficient Video DiT, and introduce a noise reinjection mechanism to balance aggressive compression with visual fidelity. On the control side, \textit{DriveLaW-Act} uses a vanilla DiT trained with a flow-matching objective to produce smooth, reliable trajectories. To harmonize optimization between generation and planning, we adopt a three-stage progressive curriculum that first learns long-horizon motion, then refines spatial detail, and finally chains video latents into the planner for stable training.

We extensively evaluate DriveLaW on nuScenes~\cite{caesar2020nuscenes} for video generation and on NAVSIM~\cite{dauner2024navsim} for trajectory planning. On nuScenes, DriveLaW achieves state-of-the-art generation quality, outperforming both pure video generators and unified gen–plan baselines. On NAVSIM, our planner attains strong closed-loop metrics without any post-training (e.g., RL) or post-processing (e.g., scorers), highlighting the strength of using video-generator representations for control. Ablations and scaling studies further substantiate our design. The main contributions of this work are as follows:
\begin{itemize}

\item We propose DriveLaW, a unified world model that shifts from parallel to chained generation and planning. We demonstrate that the latent representations learned from large-scale video generation possess superior semantic coherence and spatial structure compared to traditional Bird’s Eye View (BEV) or Vision-Language Models(VLM) features. By injecting these rich generative priors into the planner, we effectively bridge the gap between visual forecasting and action decision-making.

\item We design a specialized architecture comprising DriveLaW-Video and DriveLaW-Act. The former incorporates a novel noise reinjection mechanism to resolve structural inconsistencies and blurring in high-speed scenarios, while the latter employs a diffusion planner directly conditioned on video latents. A three-stage progressive training strategy is introduced to resolve the optimization tension between high-fidelity video synthesis and reliable trajectory generation.

\item We validate DriveLaW on standard autonomous driving benchmarks. It achieves state-of-the-art FID and FVD scores in single-view video generation on nuScenes and sets a new record for closed-loop planning metrics (PDMS) on the NAVSIM benchmark, surpassing previous world-model approaches and confirming the effectiveness of our unified paradigm without relying on post-training or auxiliary scorers.
\end{itemize}

%% file: sec/2_related_works.tex
\section{Related Work}
\label{sec:related work}

\subsection{World Models for Autonomous Driving}


World models~\cite{brooks2024video,bruce2024genie,assran2025v} aim to internalize the physical structure and dynamics of the real world into a predictive latent representation for imagination, control, and planning. 
Recent studies introduce world models into autonomous driving for scene generation~\cite{wang2024drivedreamer,zhao2025drivedreamer,gao2024vista,hassan2025gem,gao2023magicdrive,hu2023gaia,wang2025mila,gao2025magicdrive,guo2025genesis,guo2025dist,ma2024unleashing}, simulation evaluation~\cite{ni2025recondreamer,li2025omninwm,yang2025resim,yan2024street,zhou2024hugsim} and decision-making~\cite{zhang2025epona,wang2024driving,li2025drivevla,li2024enhancing,li2025end}. 
Early works~\cite{hu2023gaia,gao2023magicdrive,gao2024vista} treat this task as conditional multi-view video generation, prioritizing multi-camera geometry consistency. To more fundamentally capture the 3D spatiotemporal structure, OccWorld~\cite{zheng2024occworld} and OccSora~\cite{wang2024occsora} introduce an occupancy-centric representation.
UniScenes~\cite{li2025uniscene} and Genesis~\cite{guo2025genesis} further achieve joint generation of multi-modal signals (RGB, LiDAR, occupancy), primarily serving as data-centric generators for downstream task training.
Meanwhile, some works~\cite{yang2025resim,zhou2024hugsim,yan2024street,li2025omninwm,gao2025rad} integrate world models as simulation engines directly into autonomous driving, focusing on closed-loop evaluation and scene reconstruction.
For instance, HUGSIM~\cite{zhou2024hugsim}, ReconDreamer-R~\cite{ni2025recondreamer}, and RAD~\cite{gao2025rad} introduce 3D Gaussian Splatting to reconstruct scenes for closed-loop simulation. ReSim~\cite{yang2025resim} and OmniNWM~\cite{li2025omninwm} focus on behavioral simulation, synthesize videos conditioned on candidate trajectories, and provide reward signals to guide trajectory filtering and reinforcement learning.

However, these approaches provide only indirect guidance to the planning module via data generation and simulation, whereas recent studies aim to integrate generative modeling and decision-making within a unified world model. DrivingGPT~\cite{chen2025drivinggpt} adopts a GPT-style autoregressive policy to generate future videos and trajectories. Epona~\cite{zhang2025epona} uses an autoregressive diffusion scheme to produce videos and trajectories in parallel. VaViM/VaVAM~\cite{bartoccioni2025vavim} and DriveVLA-W0~\cite{li2025drivevla} employ a $\pi$0-like~\cite{black2024pi_0} mixture-of-transformers~\cite{liang2024mixture} architecture that autoregressively generates both modalities within a single model. FSDrive~\cite{zeng2025futuresightdrive} unifies generation and planning via visual spatio-temporal Chain-of-Thought reasoning for end-to-end driving.
While this paradigm represents a significant step forward, it typically treats video and trajectory generation as two independent output streams. This design can lead to a representation disconnect as the planning trajectory is not directly grounded in the internal features that govern video synthesis. In contrast, we argue that the rich, spatiotemporally grounded features learned by video generators constitute a powerful yet under-explored resource for planning. To bridge this gap, DriveLaW is the first to exploit mid-level latents from a video generator as planning representations, enabling more stable closed-loop driving.

\subsection{Video Generation}
Video generation has emerged as a core component of autonomous driving~\cite{lu2024wovogen,wang2024drivedreamer,wang2024driving,yang2024generalized}, underpinning vision-centric world models~\cite{hu2023gaia,hu2024drivingworld}. Beyond static scene prediction, it captures temporal continuity and rich dynamics~\cite{villegas2017decomposing,lotter2016deep,blattmann2023stable,harvey2022flexible,ho2022imagen}, enabling models to learn how agents and environments evolve over time. GAIA-1~\cite{hu2023gaia} innovatively adopts an autoregressive framework that outputs discrete image tokens to generate autonomous-driving scene videos. DriveDreamer~\cite{zhao2025drivedreamer}, Panacea~\cite{wen2024panacea}, DrivingDiffusion~\cite{li2024drivingdiffusion} and MagicDrive~\cite{gao2023magicdrive} condition diffusion models on geometric features such as Bird's Eye View (BEV) maps and 3D boxes to synthesize controllable driving scenes. MagicDrive-V2~\cite{gao2025magicdrive}, MiLA~\cite{wang2025mila}, and GAIA-2~\cite{russell2025gaia} generate long-horizon, high-fidelity driving videos via latent diffusion with structured geometric and action conditioning. TeraSim-World~\cite{wang2025terasim} and Cosmos-Drive~\cite{ren2025cosmos} build large-scale synthetic data pipelines for autonomous driving, producing diverse and controllable video samples for downstream perception training and evaluation. 
In summary, while existing methods have advanced the state of the art in visual synthesis, they predominantly treat the video generator as a renderer. We posit that the internal activations of these powerful generators encode a rich, temporally coherent understanding of scene dynamics and geometry. DriveLaW repurposes the video generator as a feature extractor and pairs it with a diffusion planner to enable end-to-end driving.

\subsection{Diffusion Policies for Autonomous Driving}
Recent work brings diffusion policies~\cite{chi2025diffusion} to autonomous driving, leveraging their strength in temporal action modeling. DiffusionDrive~\cite{liao2025diffusiondrive} introduces truncated diffusion and anchor-initialized noise to achieve real-time, multimodal trajectory planning. Diffusion Planner~\cite{zheng2025diffusion} jointly generates trajectories for the ego vehicle and surrounding agents via diffusion to model interactive driving scenes. ReCogDrive~\cite{li2025recogdrive} couples a VLM with a diffusion planner, injecting driving cognition priors into the diffusion process to enable efficient, continuous action generation. GoalFlow~\cite{xing2025goalflow} employs flow matching to generate goal-point guidance, enabling safe and stable driving. We innovatively chain a Video DiT with a diffusion planner, distill driving priors from large-scale driving videos, and inject them into a diffusion planner to enable stable closed-loop driving.

%% file: sec/3_method.tex
\section{Method}

\begin{figure*}[t] 
    \centering \includegraphics[width=\linewidth]{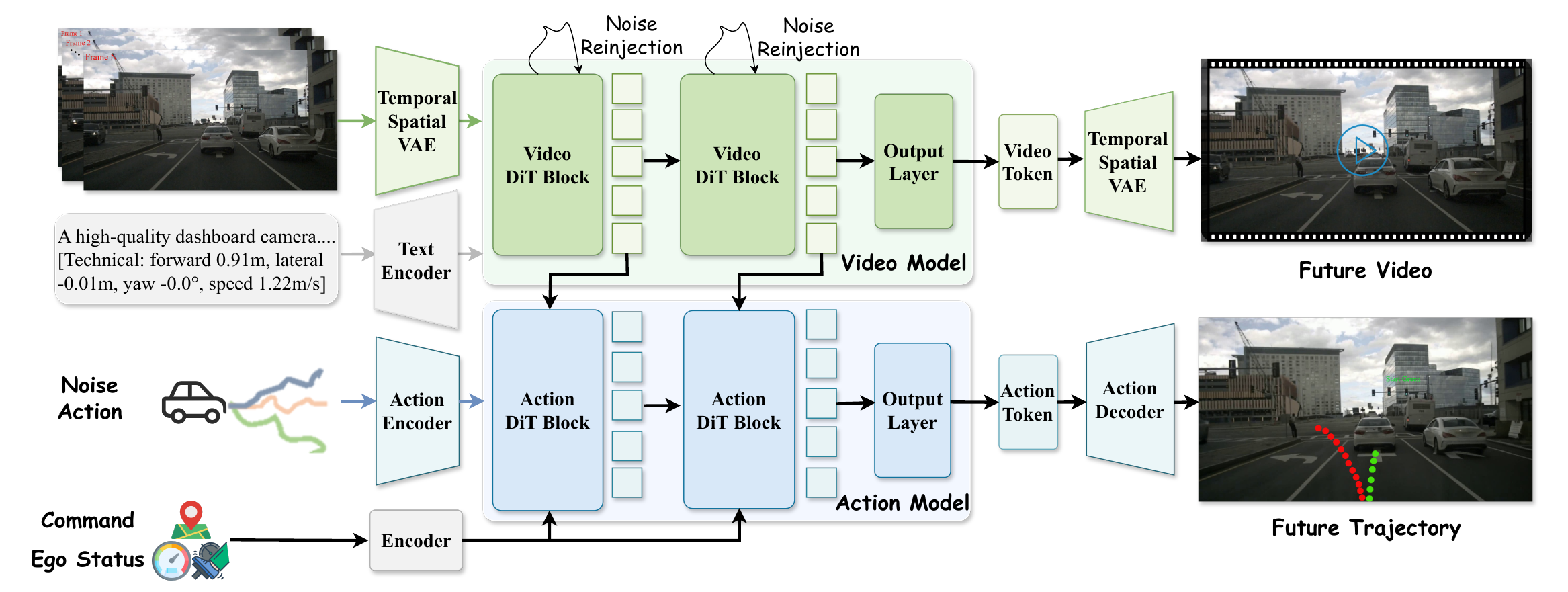} 
    \caption{\textbf{Overview of the overall architecture of DriveLaW.} The model first encodes historical observations (images, actions) into a unified latent world representation through a powerful video diffusion model. In order to improve the generation quality, we introduced the Noise Reinjection mechanism to explore and select the optimal generation path in the early stage of denoising. The denoised video latents produced by the Video DiT are then passed as conditioning signals to the action planner. Leveraging these latents, the lightweight Action DiT predicts future trajectories that are aligned with the visual scene evolution. In this chained design, the Video Model and Action Model share the same latent-space representation. }
    \label{fig:framework} 
\end{figure*}


In this section, we first investigate learning generalizable driving representations from a video generator. We then introduce DriveLaW, a world model that unifies generation and planning through a shared latent-space representation. Subsequently, we describe DriveLaW-Video, a spatiotemporal video generator, and DriveLaW-Act, a diffusion-based planner. Finally, we present a three-stage training framework that produces high-quality videos and stable trajectories.

\subsection{Learning Generalizable Driving Representations from Video Generators}
World models such as Genie~\cite{bruce2024genie} and Cosmos~\cite{ali2025world} learn real-world structure by training on large-scale video generation, and many studies~\cite{wiedemer2025video,guo2025video} indicate that video generators internalize physical regularities and act as strong zero-shot learners. In autonomous driving, real scenes provide virtually unlimited video while dense annotation is costly, so we propose to learn driving representations from large-scale driving video generation, akin to how humans acquire driving competence. Concretely, let $\mathcal{E}$ be the video encoder and $z$ the latent representation of a clip $x_0$ with $z=\mathcal{E}(x_0)$. A generic denoiser produces a latent denoising trajectory $\{z_t\}_{t\in\mathcal{T}}$ under conditioning $c$ via a single-step update
\begin{equation}
z_{t-1}=\Psi_{\theta}(z_t, t, c),
\end{equation}
where $\Psi_{\theta}$ denotes the learned denoising operator and $t$ indexes the inference schedule. We extract mid-denoising features
\begin{equation}
h_t=\phi_{\theta}(z_t),\quad t\in\mathcal{T},
\end{equation}
and select one or a small set of timesteps $t^\star$ to form the perception latent $h=h_{t^\star}$. This latent encodes driving cognition priors distilled from generation and is fed to the planning module for stable closed-loop driving.

\subsection{DriveLaW}
As illustrated in Fig.~\ref{fig:framework}, DriveLaW is a unified framework composed of a \emph{DriveLaW-Video} and a \emph{DriveLaW-Act}. The video model, e.g., LTX-Video~\cite{hacohen2024ltx}, first encodes past driving frames with a spatiotemporal VAE and encodes textual prompts with a text encoder. A stack of Video DiT blocks then performs latent-space denoising, and the VAE decoder reconstructs the video. Concurrently, action noise, ego status, and high-level commands are encoded and fed into the action model. Video latents from the Video DiT serve as conditioning signals, guiding the Action DiT to output the final trajectory. The Video DiT and Action DiT are chained and trained to learn driving representations from large-scale video generation, providing a shared basis for planning.

\subsection{DriveLaW-Video: Spatiotemporal World Generator}
\paragraph{Spatiotemporal VAE.}

We employ a high-compression spatiotemporal VAE~\cite{hacohen2024ltx} to efficiently model long-horizon driving scenarios. The VAE encodes each video clip into a causal latent space with $32\times32\times8$ spatial-temporal resolution and 128 channels, achieving a compression ratio of $1{:}192$ (pixel-to-token ratio $1{:}8192$). This is significantly more compact than typical $1{:}48$~\cite{polyak2024movie,lei2023pyramidflow} or $1{:}96$~\cite{kong2024hunyuanvideo,yang2024cogvideox} compressions, enabling longer prediction horizons under the same computational budget—critical for modeling long-term dependencies like traffic light changes and vehicle dynamics. The encoder uses 3D causal convolutions to ensure each timestep depends only on past and current frames, preventing temporal information leakage.


Unlike conventional pipelines that complete all reverse diffusion steps in the latent space before a single decoding pass, we employ a hybrid approach. We decode at a late stage of the rectified-flow schedule ($t = t_1$) and perform a final refinement in pixel space:
\begin{equation}
x_0 = D(z_{t_1}, t_1),
\end{equation}
where
\begin{equation}
z_{t_1} = (1 - t_1) z_0 + t_1 \epsilon, \quad \epsilon \sim \mathcal{N}(0,\mathbf{I}).
\end{equation}
Here, $D$ denotes a time-conditioned denoising decoder trained with pixel-space losses. Since $D$ maps latents to pixels, it is used only for the final step. Performing the last step in pixel space recovers high-frequency details, e.g., highlights, dynamic shadows, fine road textures, without an extra super-resolution module and adds minimal overhead.

\paragraph{Video Transformer Architecture.}

After compression to the high-dimensional latent space, the serialized spatiotemporal tokens are processed by a three-dimensional Transformer adapted from PixArt-$\alpha$~\cite{chen2023pixart} for full spatiotemporal modeling. Each block uses self-attention~\cite{vaswani2017attention} for global spatiotemporal modeling and cross-attention to integrate task-specific conditions (navigation commands, visual cues). We apply RMSNorm~\cite{zhang2019root} to queries and keys for attention stability. To enhance spatiotemporal consistency under different resolutions and frame rates, we use Rotary Positional Embedding~\cite{su2024roformer} with normalized fractional coordinates, reducing spatial drift in long-horizon predictions.


To align video generation with realistic driving dynamics, we introduce a motion-conditioned prompting mechanism that converts recent ego-vehicle kinematics into structured natural language instructions rather than using a dedicated motion encoder. This approach leverages pre-trained text-to-video architectures directly, provides interpretable control, unifies static and dynamic conditioning, and improves cross-dataset generalization by avoiding numeric encodings tied to data-specific scales.


\paragraph{Noise Reinjection.}

\begin{figure}[t] 
    \centering \includegraphics[width=\linewidth]{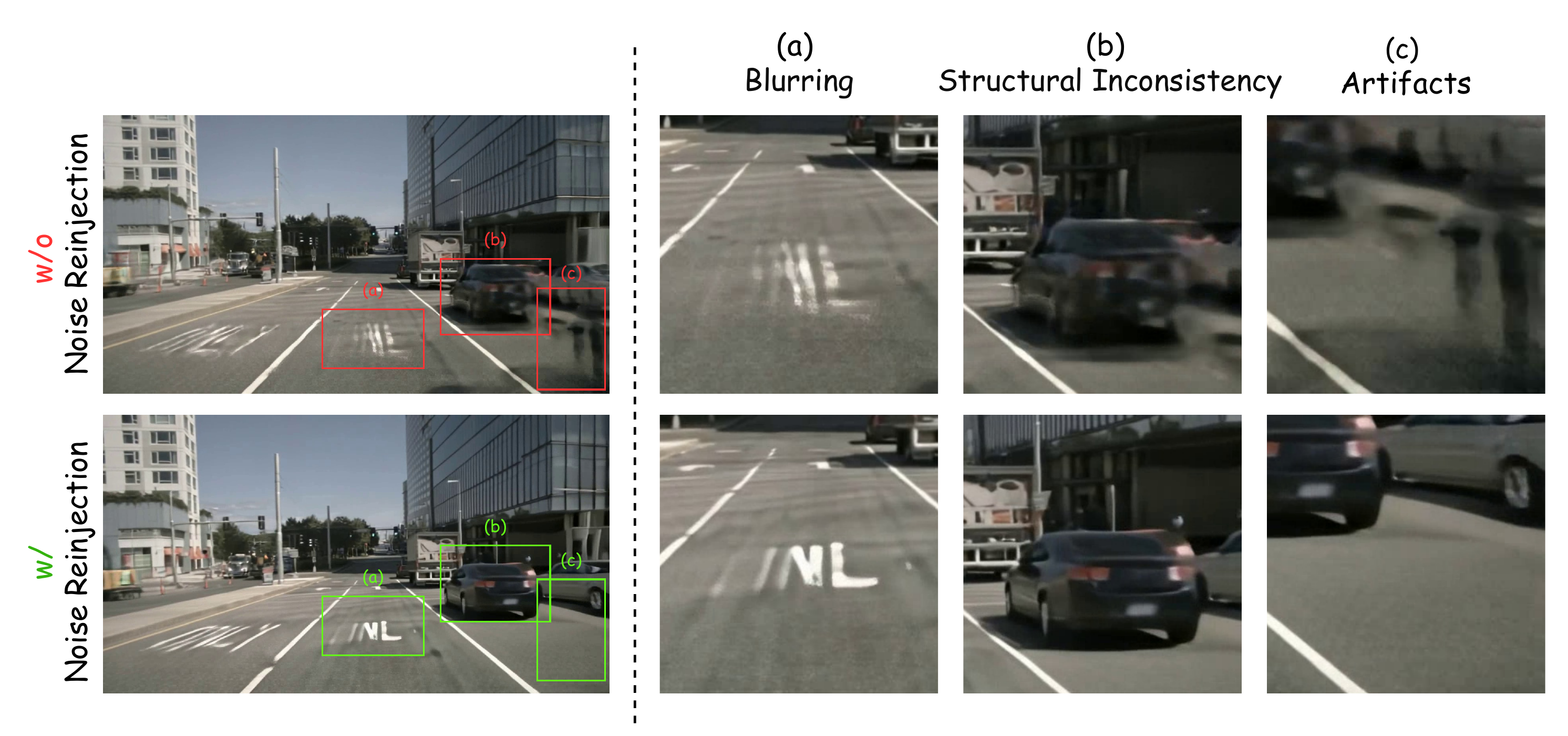} 
    \caption{\textbf{Restoring Structural and Temporal Consistency via Noise Reinjection.} This comparison highlights the impact of our method. The baseline generation shows significant degradation, including (a) blurring, (b) structural inconsistency, and (c) artifacts. By integrating noise reinjection, our model preserves sharp details, maintains object structures, and produces clean, artifact-free frames, demonstrating a crucial improvement in video quality.}
    \label{fig:renoise} 
\end{figure}

In high-speed driving video generation, long-range motion and large displacements often cause perceptual degradation: boundaries are over-smoothed, fine textures fade, and blur and ghosting accumulate, undermining the structural consistency of vehicles, lane markings, and distant backgrounds. To mitigate this, we adapt the principle of iterative refinement found in works like DiffuseSlide~\cite{hwang2025diffuseslide}. Unlike methods that apply noise globally, we introduce a more targeted strategy. Our approach selectively re-injects noise into high-frequency regions before each main denoising step, compelling the model to actively regenerate details rather than smooth them over.

At each denoising step $t$ with current latent $L_t$, we first identify regions likely to contain high-frequency details. To do this stably, we perform an initial prediction of the clean latent $\hat{L}_0 = \Psi_	\theta(L_t, t)$. Crucially, we then decode this latent into a temporary pixel-space image using the VAE decoder, $\hat{I}_0 = D(\hat{L}_0)$, and convert each frame to its grayscale representation, $G_f$. This entire process of computing the high-frequency mask occurs in the pixel domain for maximum fidelity. We then apply a discrete Laplacian kernel $K_L$ to obtain a response map $H_f = |G_f * K_L|$, compute an adaptive threshold $	au = \beta \cdot \mathrm{std}(H_f)$, and define the high-frequency mask:
\begin{equation}
M_f(x, y) =
\begin{cases}
1, & H_f(x, y) > 	au \\
0, & 	\text{otherwise}.
\end{cases}
\end{equation}
To apply this mask back in the latent space, we downsample $M_{f, 	\text{pixel}}$ from the image resolution ($H \times W$) to the latent resolution ($h \times w$) using nearest-neighbor interpolation, resulting in the final latent-space mask $M$. Next, we create a selectively perturbed latent $L'_t$ by injecting a small amount of controlled noise only in the masked regions:
\begin{equation}
L'_t = L_t + \sigma'_t \cdot M \odot \varepsilon_t, \quad \varepsilon_t \sim \mathcal{N}(0,\mathbf{I}),
\end{equation}
where $\sigma'_t$ is a manually-tuned noise strength for the reinjection step, and $M$ is the high-frequency mask.

Finally, this perturbed latent $L'_t$ is used as the input to the full Transformer-based denoising operator $\Psi_	\theta$ to compute the latent for the next step, $L_{t-\Delta t}$. This forces the model to leverage its powerful generative prior to "inpaint" the noisy regions with plausible high-frequency details consistent with the rest of the scene. As shown in Fig.~\ref{fig:renoise}, this targeted approach restores sharpness and texture to dynamic objects and road markings while preserving the natural smoothness of areas like the sky, achieving a better balance between detail restoration and artifact suppression.

\subsection{DriveLaW-Act: A Diffusion-Based Planner}

Inspired by~\cite{zheng2025diffusion,liao2025diffusiondrive,li2025recogdrive}, we adopt a diffusion-based planner that generates continuous, temporally smooth trajectories. Specifically, we encode a sampled noise action and the driving context as follows. Concretely, we encode a noised action $a_t$, ego status $s_t$, and high-level command $g_t$ using an action encoder and a context encoder respectively:
\begin{equation}
\begin{aligned}
h_{\mathrm{act}} &= E_{\mathrm{act}}(a_t), \qquad
h_{\mathrm{ctx}} = E_{\mathrm{ctx}}([s_t; g_t]),
\end{aligned}
\end{equation}
where $a_t = (1 - t)a_0 + t\epsilon$ and $\epsilon \sim \mathcal{N}(0, \mathbf{I})$. Meanwhile, during the first denoising step of the Video DiT, the latent features from each Transformer block are cached as $\{f_1, f_2, \dots, f_B\}$.
For every step in the flow-matching process, the Action DiT (denoted as $f_\theta$) takes the encoded noise action $h_{\mathrm{act}}$ and the continuous time $t$ as input, conditioned on both the context embedding $h_{\mathrm{ctx}}$ and the cached video features $\{f_i\}$:
\begin{equation}
f_\theta(a_t, t) = \mathrm{DiT}_{\mathrm{act}}\big([\,h_{\mathrm{act}};\, t\,] \,\big|\, h_{\mathrm{ctx}}, \{f_i\}_{i=1}^{B}\big),
\end{equation}
where $t$ denotes the continuous timestep.

We train the planner using a flow-matching~\cite{lipman2022flow} objective that aligns the model's predicted output $f_\theta$ with the target flow:
\begin{equation}
\mathcal{L}_{\mathrm{FM}} = \mathbb{E}_{t,\,a_0,\,\epsilon}\Big[\big\|f_\theta(a_t, t) - (a_0 - \epsilon)\big\|_2^2\Big].
\end{equation}
This encourages smooth, stable trajectory generation consistent with the learned driving dynamics.

\subsection{Three-Stage Progressive Training}

To produce high-quality, stable driving videos while furnishing the planner with strong representations, we adopt a three-stage progressive training scheme.

In the first stage, we focus on learning robust motion patterns by training on longer clips at a reduced spatial resolution, $740 \times 352 \times 121$ (width $\times$ height $\times$ frames). This configuration prioritizes temporal span over spatial detail, enabling the model to learn smooth, continuous driving behaviors such as lane keeping, turning, and speed variations. Because memory in video diffusion scales with both spatial and temporal extents, lowering the resolution allows for processing more frames, which is crucial for modeling long-horizon scenarios.

Subsequently, we switch to higher spatial resolution but shorter clips, $1280 \times 704 \times 25$, to further enhance the visual quality and fine-grained details, such as lane markings, surrounding vehicles, and environmental textures. In this phase, the larger spatial dimensions with fewer frames allocate capacity to spatial fidelity, while preserving the temporal coherence established in the first stage.

Finally, building on this strong video generator that learns physically grounded driving dynamics, we condition DriveLaW-Act on latent features from DriveLaW-Video and train it for trajectory planning. This third stage couples generation and planning by using video latents as compact perception for the planner. The three-stage curriculum equips DriveLaW with high-fidelity video synthesis and reliable, stable trajectory planning.

%% file: sec/4_experiment.tex
\section{Experiment}

\begin{table*}[t]
\centering
\caption{\textbf{Quantitative evaluation of video generation on the NuScenes validation set.} Our method outperforms prior single-view state-of-the-art methods in generation quality.}
\label{tab:prediction_fidelity}
\resizebox{\textwidth}{!}{%
\begin{tabular}{@{}lccccccc@{}}
\toprule
\textbf{Metric} & DriveGAN~\cite{kim2021drivegan} & DriveDreamer~\cite{wang2024drivedreamer} & DrivingGPT~\cite{chen2025drivinggpt} & DriveWorld~\cite{min2024driveworld} & Vista~\cite{gao2024vista} & Epona~\cite{zhang2025epona} & \textbf{DriveLaW (Ours)} \\
\midrule
\textbf{FID} $\downarrow$ & 73.4 & 52.6 & 12.8 & 7.4 & 6.9 & 7.5 & \textbf{4.6} \\
\textbf{FVD} $\downarrow$ & 502.3 & 452.0 & 142.6 & 90.9 & 89.4 & 82.8 & \textbf{81.3} \\
\bottomrule
\end{tabular}%
}
\end{table*}


\begin{table*}[t!]
    \centering
    \caption{\textbf{Performance comparison on NAVSIM \textit{Navtest} using closed-loop metrics.} Methods are grouped by whether they employ an explicit world model: \textit{Traditional End-to-End Methods} and \textit{World Model Methods}. $^{\dagger}$ denotes methods trained with the same flow-matching objective.}
    \setlength{\tabcolsep}{4pt}
    \begin{tabular}{@{}l|c|cc|cc|ccc|cc@{}}
        \toprule
        \textbf{Method} & \textbf{Ref} & \textbf{Image} & \textbf{Lidar} & \textbf{NC$\uparrow$} & \textbf{DAC$\uparrow$} & \textbf{TTC$\uparrow$} & \textbf{Comf.$\uparrow$} & \textbf{EP$\uparrow$} & \textbf{PDMS$\uparrow$} \\
        \midrule
        Constant Velocity & - & & & 68.0 & 57.8 & 50.0 & \textbf{100} & 19.4 & \cellcolor{gray!30} 20.6 \\
        Ego Status MLP & arXiv'23 & &  & 93.0 & 77.3 & 83.6 & \textbf{100} & 62.8 & \cellcolor{gray!30} 65.6 \\     
        \midrule
        \multicolumn{8}{@{}l}{\raggedright \textit{Traditional End-to-End Methods}} \\
        VADv2-$\mathcal{V}_{\text{8192}}$~\citep{chen2024vadv2} & arXiv'24 & \checkmark &  &  97.2 & 89.1 & 91.6 & \textbf{100} & 76.0 &\cellcolor{gray!30}  80.9 \\
        UniAD~\citep{hu2023planning} & CVPR'23 & \checkmark &    & 97.8 & 91.9 & 92.9 & \textbf{100} & 78.8 & \cellcolor{gray!30} 83.4 \\
        TransFuser~\citep{chitta2022transfuser} & TPAMI'23 & \checkmark & \checkmark   & 97.7 & 92.8 & 92.8 & \textbf{100} & 79.2 & \cellcolor{gray!30} 84.0 \\
        PARA-Drive~\citep{weng2024drive} & CVPR'24 & \checkmark &  & 97.9 & 92.4 & 93.0 & 99.8 & 79.3 & \cellcolor{gray!30} 84.0 \\
        ReCogDrive-IL~\cite{li2025recogdrive} & arXiv'25 & \checkmark &  &  98.1 & 94.7 & 94.2 & \textbf{100} & 80.9 & \cellcolor{gray!30} 86.5 \\
        DiffusionDrive~\cite{liao2025diffusiondrive}  & CVPR'25 & \checkmark & \checkmark & 98.2 & 96.2 & 94.7 & \textbf{100} & \textbf{82.2} & \cellcolor{gray!30} 88.1 \\
        \midrule
        \multicolumn{8}{@{}l}{\raggedright \textit{World Model Methods}} \\
        DrivingGPT~\cite{chen2025drivinggpt} & arXiv'24 & \checkmark &  &  98.9 & 90.7 & 94.9 & 95.6 & 79.7 & \cellcolor{gray!30} 82.4 \\ 
        LAW~\cite{li2024enhancing} & ICLR'25 & \checkmark &  &  96.4 & 95.4 & 88.7 & 99.9 & 81.7 & \cellcolor{gray!30} 84.6 \\ 
        Epona~\cite{zhang2025epona}  & ICCV'25 & \checkmark &  &  97.9 & 95.1 & 93.8 & 99.9 & 80.4 & \cellcolor{gray!30} 86.2 \\
        Resim~\cite{yang2025resim} & NeurIPS'25 & \checkmark &  &  -- & -- & -- & -- & -- & \cellcolor{gray!30} 86.6 \\
        WoTE~\citep{li2025end} & ICCV'25 & \checkmark & \checkmark & 98.5 & 96.8 & 94.9 & 99.9 & 81.9 & \cellcolor{gray!30} 88.3 \\
        DriveVLA-W0$^{\dagger}$~\citep{li2025drivevla} & arXiv'25 & \checkmark &  & 98.4 & 95.3 & 95.2 & \textbf{100} & 80.9 & \cellcolor{gray!30} 87.2 \\
        PWM~\cite{zhao2025forecasting} & NeurIPS'25 & \checkmark &  &  98.6 & 95.9 & 95.4 & \textbf{100} & 81.8 & \cellcolor{gray!30} 88.1 \\
        \textbf{DriveLaW(Ours)} & - & \checkmark &  & \textbf{99.0} & \textbf{97.1} & \textbf{96.7} & \textbf{100} & 81.3 & \cellcolor{gray!30} \textbf{89.1} \\
        \bottomrule
    \end{tabular}
    \label{tab:comparison_modified}
\end{table*}

\begin{table}[h]
\centering
\caption{\textbf{Planning performance on NuScenes.} We report L2 displacement error and collision rate at 1s, 2s, 3s, and averaged.}
\setlength{\tabcolsep}{4pt}
\scalebox{0.9}{ 
\begin{tabular}{@{}l|cccc|cccc@{}} 
\toprule
\textbf{Method} & \multicolumn{4}{c|}{\textbf{L2 (m) $\downarrow$}} & \multicolumn{4}{c}{\textbf{Collision (\%) $\downarrow$}} \\ 
\cmidrule(lr){2-5} \cmidrule(lr){6-9}
                & 1s & 2s & 3s & Avg. & 1s & 2s & 3s & Avg. \\
\midrule
Epona~\cite{zhang2025epona}           & 0.61 & 1.17 & 1.98 & 1.25 & \textbf{0.01} & 0.22 & 0.85 & 0.36 \\
\textbf{DriveLaW} & \textbf{0.44} & \textbf{1.10} & \textbf{1.91} & \textbf{1.15} & 0.15 & \textbf{0.10} & \textbf{0.48} & \textbf{0.24} \\
\bottomrule
\end{tabular}
}
\label{tab:realworld}
\end{table}

\subsection{Experimental Setup}

\paragraph{Implementation Details.}
\label{sec:implementation_details}

Our DriveLaW model consists of a 2B video DiT, initialized from LTX-Video~\cite{hacohen2024ltx} pretrained weights, and a 133M diffusion planner for trajectory planning. We enable DriveLaW to acquire both video generation and planning capabilities through a three-stage training pipeline consisting of video pretraining followed by action fine-tuning. In the video pretraining stage, we train the Video DiT~\cite{peebles2023scalable} on $8\,\mathrm{Hz}$ frames from nuScenes~\cite{caesar2020nuscenes} and NuPlan~\cite{caesar2021nuplan} with a two-stage curriculum. We first establish temporal coherence by training on low-resolution, long-duration clips ($740 \times 352 	\times 121$ frames). Then we fine-tune the model on high-resolution, shorter clips ($1280 \times 704 \times 25$ frames). First two stages are trained for 30k iterations with a batch size of 4, a learning rate of $1\times10^{-5}$, and a weight decay of $5\times10^{-2}$. We use flow matching~\cite{lipman2022flow} with token-wise uniform $\sigma\in[0,1]$, so each latent token interpolates independently between data and noise. In the trajectory fine-tuning stage, we feed the past four camera frames and supervise $2\,\mathrm{Hz}$ trajectory points over the next 4s, updating both the Video DiT and the Planning DiT. We use a batch size of 192 for 44k steps with a fixed learning rate of $3\times10^{-5}$ and a weight decay of $1\times10^{-5}$. At inference, we use 30 sampling steps for video generation and 5 steps for trajectory planning.



\paragraph{Dataset and Metrics.}

Following the previous autonomous driving experimental protocol~\cite{zhang2025epona,li2025drivevla,zhao2025forecasting}, we train our model on nuPlan and nuScenes, and subsequently evaluate it on nuScenes for video generation and on NAVSIM~\cite{dauner2024navsim} for trajectory planning. The nuScenes dataset comprises 1,000 driving scenes collected in Boston and Singapore, featuring multi-sensor data from cameras and LiDAR, with 850 scenes used for training and validation and the remaining 150 reserved for testing. NuPlan, the first large-scale planning dataset for autonomous driving, provides 1,200 hours of human driving data from four cities. NAVSIM~\cite{dauner2024navsim} further builds on OpenScene~\citep{peng2023openscene}, a redistribution of nuPlan, offering a planning-oriented benchmark for trajectory prediction. It is divided into two subsets: \textit{Navtrain}, containing 103k samples, and \textit{Navtest}, containing 12k samples. We utilize 8 Hz driving camera videos from nuScenes and NuPlan for video training, and 2 Hz camera data from NAVSIM for trajectory prediction.


For video generation quality evaluation on the nuScenes validation set, we use FVD~\cite{unterthiner2018towards} and FID~\cite{heusel2017gans} for assessment. For planning evaluation on NAVSIM~\cite{dauner2024navsim}, we use the Predictive Driver Model Score (PDMS), which combines penalties for no-at-fault collisions (NC) and drivable-area compliance (DAC) with weighted measures of ego progress (EP), time-to-collision (TTC), and comfort (Comf.) to evaluate overall safety, compliance, and efficiency.

\subsection{Main Results}
\paragraph{Quantitative experiments on video generation.}
Tab.~\ref{tab:prediction_fidelity} reports the quantitative results of the generative evaluation on nuScenes. In both metrics, DriveLaW surpassing all previous single-view approaches with state-of-the-art performance of 4.6 FID and 81.3 FVD. Results demonstrate the effectiveness of our method for high-fidelity driving video generation.

\paragraph{Quantitative experiments on motion planning.}

Tab.~\ref{tab:comparison_modified} reports NAVSIM results. DriveLaW attains a PDMS of 89.1, setting a new state of the art without any post-training such as reinforcement learning or post-processing such as learned scorers. It surpasses traditional end-to-end planners including DiffusionDrive~\cite{liao2025diffusiondrive}, which fuses camera and LiDAR, and ReCogDrive~\cite{li2025recogdrive}, which relies on Vision-Language Models. Compared with world-model methods, DriveLaW improves over Epona~\cite{zhang2025epona} by 2.9 PDMS, where Epona adopts a parallel generation–planning design, and over DriveVLA-W0~\cite{li2025drivevla} and PWM~\cite{zhao2025forecasting} by 1.9 and 1.0 PDMS respectively, which use VLMs and world-model supervision, demonstrating the effectiveness of chaining generation with planning.

\paragraph{Additional performance on motion planning.}
As an additional open-loop evaluation, we report planning performance on the NuScenes validation set in Tab.~\ref{tab:realworld}. DriveLaW achieves lower L2 errors and collision rates compared to Epona~\cite{zhang2025epona}, demonstrating robust generalization and safety on real-world data.

\begin{figure*}[t] 
    \centering \includegraphics[width=\linewidth]{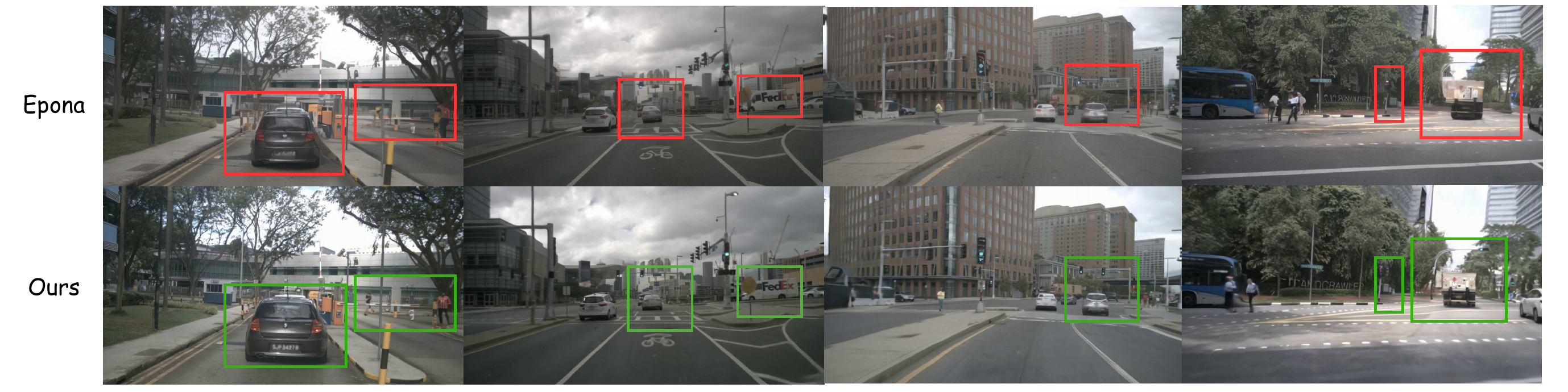} 
    \caption{\textbf{Qualitative Comparison} with state-of-the-art driving world model. We compare DriveLaW with Epona~\cite{zhang2025epona} on nuScenes validation set. DriveLaW generates videos with (1) clearer vehicle details and more stable structural integrity, (2) well-preserved pedestrian shapes that remain easily identifiable, and (3) correct recognition and maintenance of inconspicuous objects (e.g., the yellow van), demonstrating superior visual quality, subject preservation, and semantic understanding.}
    \label{fig:contrast} 
\end{figure*}

\paragraph{Qualitative Results.}
 Fig.~\ref{fig:contrast} presents qualitative comparisons between our DriveLaW and the current state-of-the-art open-source driving world model Epona~\cite{zhang2025epona}. As shown in the leftmost pair of images, vehicles in videos generated by Epona exhibit noticeable visual artifacts, whereas DriveLaW produces results with clearer vehicle details and more stable structural integrity. In Epona’s outputs, pedestrian figures nearly degrade to the point of being unrecognizable, while DriveLaW preserves complete shapes that remain easily identifiable. Additionally, in the case of the visually inconspicuous yellow van in the scene, Epona misclassifies it as a different object, whereas DriveLaW correctly recognizes and maintains its appearance and spatial position. These results demonstrate that DriveLaW excels in visual quality, subject preservation, and semantic world understanding.


\subsection{Ablation Study}


\paragraph{Planning Gains from Scaling Video Generator Pretraining.}
As shown in Tab.~\ref{tab:navsim_scaling_law}, increasing pretraining samples for the video generator consistently boosts DriveLaW’s closed-loop performance on NAVSIM. A fully pretrained generator delivers a +3.2 PDMS improvement over a generator without driving-domain pretraining, indicating that larger corpora deepen the model’s grasp of driving physics and translate into stronger planning, exhibiting a clear scaling law.

\begin{table}[t]
    \centering
    \caption{\textbf{Scaling video pretraining improves planning on NAVSIM \textit{Navtest}.} Rows vary the number of video pretraining samples used before fine-tuning the diffusion planner on NAVSIM.}
    \setlength{\tabcolsep}{3.5pt}\footnotesize
    \begin{tabular*}{\columnwidth}{@{\extracolsep{\fill}} l | c c c c c | c @{\hspace{\tabcolsep}}}
        \toprule
        \textbf{Video P.T. Size} & \textbf{NC$\uparrow$} & \textbf{DAC$\uparrow$} & \textbf{TTC$\uparrow$} & \textbf{Comf.$\uparrow$} & \textbf{EP$\uparrow$} & \textbf{PDMS$\uparrow$} \\
        \midrule
        0 (scratch) & 98.2 & 93.8 & 94.1 &  99.9 & 80.8 & \cellcolor{gray!30}85.9 \\
        76k         & 98.7 & 94.7 & 95.3 & 99.9 & 80.8 & \cellcolor{gray!30}87.0 \\
        3.8M        & 98.6 & 95.8 & 94.8 & \textbf{100} & \textbf{82.2} & \cellcolor{gray!30}87.8 \\
        7.6M        & \textbf{99.0} & \textbf{97.1} & \textbf{96.7} & \textbf{100} & 81.3 & \cellcolor{gray!30} \textbf{89.1}  \\
        \bottomrule
    \end{tabular*}
    \label{tab:navsim_scaling_law}
\end{table}


\paragraph{Comparison of Driving Representations on NAVSIM \textit{Navtest}.}
Tab.~\ref{tab:ablation_repr} reports results with different driving representations. Under the same diffusion based planner, video generator latent features improve PDMS by 5.0 over BEV features and by 2.6 over VLM hidden states, demonstrating the effectiveness of this representation.

\begin{table}[t]
    \centering
    \caption{\textbf{Representation ablation on NAVSIM \textit{Navtest}.} We compare BEV features, VLM hidden states, and video latents as diffusion condition.}
    \label{tab:ablation_repr}
    \setlength{\tabcolsep}{3.5pt}\footnotesize
    \begin{tabular*}{\columnwidth}{@{\extracolsep{\fill}} l | c c c c c | c @{\hspace{\tabcolsep}}}
        \toprule
        \textbf{Representation} & \textbf{NC$\uparrow$} & \textbf{DAC$\uparrow$} & \textbf{TTC$\uparrow$} & \textbf{Comf.$\uparrow$} & \textbf{EP$\uparrow$} & \textbf{PDMS$\uparrow$} \\
        \midrule
        BEV Features            & 97.6 & 93.0 & 92.9 & \textbf{100} & 79.1 & \cellcolor{gray!30}84.1 \\
        VLM Hidden State        & 98.1 & 94.7 & 94.2 & \textbf{100} & 80.9 & \cellcolor{gray!30}86.5 \\
        Video Latents    & \textbf{99.0} & \textbf{97.1} & \textbf{96.7} & \textbf{100} & \textbf{81.3} & \cellcolor{gray!30} \textbf{89.1} \\
        \bottomrule
    \end{tabular*}
\end{table}

\paragraph{Ablation on the Video Denoising Step for Action DiT Conditioning.}
As shown in Tab.~\ref{tab:ablation_denoise_step}, we ablate which video denoising step provides the latent condition to the Action DiT. Conditioning on latents from early denoising steps yields stronger planning, while latents from later steps perform worse. This occurs because raw pixel-format videos frequently contain redundant, non-essential information, which can hinder the effectiveness of decision-making.

\begin{table}[t]
    \centering
    \caption{\textbf{Which video denoising step feeds the Action DiT.} We evaluate planning when conditioning on video latents taken from different diffusion denoising steps.}
    \label{tab:ablation_denoise_step}
    \setlength{\tabcolsep}{3.5pt}\footnotesize
    \begin{tabular*}{\columnwidth}{@{\extracolsep{\fill}} l | c c c c c | c @{\hspace{\tabcolsep}}}
        \toprule
        \textbf{Video Denoise Step} & \textbf{NC$\uparrow$} & \textbf{DAC$\uparrow$} & \textbf{TTC$\uparrow$} & \textbf{Comf.$\uparrow$} & \textbf{EP$\uparrow$} & \textbf{PDMS$\uparrow$} \\
        \midrule
        $t=1$         &  99.0 & \textbf{97.1} & \textbf{96.7} & \textbf{100} & 81.3 & \cellcolor{gray!30} \textbf{89.1} \\
        $t=5$            & \textbf{99.2} & 93.7 & 95.6 & \textbf{100} & \textbf{81.8} & \cellcolor{gray!30}86.9 \\
        $t=10$          & 81.7 & 63.4 & 67.6 & 0 & 15.4 & \cellcolor{gray!30}23.2 \\
       \bottomrule
    \end{tabular*}
\end{table}

\paragraph{Effect of Training Strategy.}
Tab.~\ref{tab:method_comparison} presents ablation results of different training strategies. Removing the first stage results in comparable FID but a much higher FVD, indicating a notable loss of temporal coherence due to the absence of long-horizon motion modeling. Omitting the second stage preserves temporal stability but slightly degrades spatial detail, as reflected in a moderate FVD increase. The complete multi-stage training strategy achieves the best balance, yielding both the lowest FID and FVD, confirming that each stage plays a complementary role in ensuring high-quality and temporally consistent driving video generation.

\begin{table}[t]
\centering
\caption{\textbf{Comparison of different training strategies with FID and FVD scores.}}
\setlength{\tabcolsep}{6pt} 
\begin{tabular}{l | cc}
\hline
\textbf{Methods} & \textbf{FID$\downarrow$} & \textbf{FVD$\downarrow$} \\
\toprule
w/o First Stage Training   & 5.0 & 109.3 \\
w/o Second Stage Training  & 5.0 & 93.2 \\
\textbf{Ours}              & \bfseries 4.6 & \bfseries 81.3 \\
\bottomrule
\end{tabular}
\label{tab:method_comparison}
\end{table}

%% file: sec/5_conclusion.tex
\section{Conclusion}
In this work, we propose DriveLaW, a unified latent world model that addresses the long-standing disconnect between video generation and motion planning in autonomous driving. We first introduced DriveLaW-Video, a spatiotemporal generation module enhanced with a Noise Reinjection mechanism to ensure high-fidelity, temporally consistent video synthesis. Building on this foundation, we designed DriveLaW-Act, a diffusion-based planner that leverages video latents to generate smooth and reliable trajectories. To further harmonize optimization between generation and planning, we adopted a three-stage progressive training strategy. Extensive experiments on nuScenes and NAVSIM benchmarks demonstrate that DriveLaW achieves state-of-the-art performance, validating the effectiveness of unifying driving world generation and decision-making through a shared latent representation for next-generation end-to-end autonomous driving systems.

%% file: sec/X_suppl.tex
\clearpage

\appendix


\section{More Implementation Details}
In this section, we provide additional implementation specifics for the core components of DriveLaW, including the video generation backbone (DriveLaW-Video), the trajectory planning module (DriveLaW-Act), and the motion-conditioned prompting mechanism. 

\subsection{DriveLaW-Video: Video Generation Backbone}
DriveLaW-Video adopts a diffusion-based architecture optimized for high-compression spatiotemporal encoding and efficient chained generation with the downstream planner. This design balances computational efficiency and generation quality, enabling long-horizon driving scenario synthesis under practical hardware constraints.

\paragraph{Spatiotempora VAE and Compression Optimization.}
The Video-VAE serves as the core spatiotemporal compression module, applying a $32 \times 32 \times 8$ spatial–temporal downsampling with 128 output channels. This configuration achieves a total compression ratio of 1:192 (pixels-to-tokens ratio of 1:8192), approximately twice the compression rate of common text-to-video pipelines. To enable such aggressive compression without compromising generation fidelity, we introduce the following architectural and training modifications:
\begin{itemize}
    \item \textbf{Causal 3D Encoder}: Ensures each temporal step depends only on current and past frames, preserving the autoregressive consistency critical for driving prediction tasks.
    \item \textbf{Hybrid Decoding Strategy}: Instead of completing all denoising steps in the latent space, the final rectified-flow step ($t_1$) is executed by the VAE decoder directly in the pixel space. This design recovers high-frequency details (e.g., road texture, reflections, traffic signs) without requiring a separate super-resolution stage.
    \item \textbf{Reconstruction GAN}: The discriminator receives paired real–reconstructed samples and focuses on fine-grained detail differences. This improves training stability and perceptual quality under high compression.
    \item \textbf{Multi-layer Noise Injection}: Introduces per-channel learned stochasticity in the decoder, enhancing the diversity of synthesized textures.
    \item \textbf{Uniform Log-variance Across Channels}: Ensures balanced KL regularization and avoids underutilized latent dimensions, improving the efficiency of the latent space.
    \item \textbf{Video-DWT Loss}: Complements MSE and perceptual losses by explicitly penalizing high-frequency errors across eight 3D wavelet sub-bands, strengthening the preservation of structural details.
\end{itemize}

\paragraph{Video Transformer Backbone.}
The diffusion backbone adopts a 3D Transformer architecture adapted from PixArt-$\alpha$, with $28$ self--cross attention blocks, a hidden size of $2048$, a feed-forward expansion factor of $\times 4$, and RMSNorm normalization in place of LayerNorm for better stability.

 To maintain spatial–temporal consistency across varying resolutions and durations, we employ normalized fractional Rotary Positional Embeddings (RoPE) computed with exponential frequency spacing. Unlike patchifier-based designs (e.g., $2 \times 2 \times 1$ patch size), tokens are serialized directly from the VAE latents at a $1 \times 1 \times 1$ granularity, eliminating redundant patchification operations and preserving geometric consistency.

\subsection{DriveLaW-Act: Trajectory Planning Module}
DriveLaW-Act is implemented as a lightweight diffusion planner (133M parameters) that is tightly integrated with the DriveLaW-Video backbone. Its key design details are as follows:

\paragraph{Input Conditioning.}
The planner is directly conditioned on cached Video-DiT latents from the first denoising step. These latents encode rich scene information, including current geometry and agent dynamics, and serve as keys in the planner’s cross-attention mechanism, paired with the trajectory noise input. Additionally, the planner receives structured context embeddings, including: ego-vehicle kinematics and navigation commands.

\paragraph{Training and Inference.}
The planner is trained with a flow-matching objective to generate smooth, physically consistent trajectories. It predicts continuous $(x, y, \theta)$ positions at a sampling rate of 2~Hz over a 4~s planning horizon. During inference, the planner operates purely in the latent space without requiring video decoding, significantly reducing computational overhead. Notably, gradient isolation between the video generation and planning modules is preserved during training, ensuring stable optimization of each component.

\subsection{Motion-Conditioned Prompting Mechanism}
To align video generation with realistic driving dynamics, we design a structured motion-conditioned prompting mechanism that unifies dynamic ego-state information and static scene context into interpretable text guidance for the Video-DiT.

\paragraph{Prompt Construction Logic.}
Ego-state numerical variables (speed, steering angle, displacement) are first discretized into semantic bins (e.g., "low speed", "steady motion", "turning left/right"). These semantic labels are integrated into a fixed prompt template, which also includes technical numerical grounding to ensure precise control. The template is defined as follows:
\begin{quote}
A high-quality, photorealistic dashboard camera view of autonomous driving. Based on the past $T_h$ seconds, predict and generate the next $T_p$ seconds of realistic driving continuation, moving at [speed bin] with [motion descriptor], smoothly continue for the next $T_p$ seconds. Maintain temporal consistency, stable camera perspective, natural motion flow without jitter or artifacts, clear details, and realistic physics. [Technical: forward $\Delta x$ m, lateral $\Delta y$ m, yaw $\Delta \theta^\circ$, speed $v$ m/s]
\end{quote}

The text prompt is encoded by the frozen T5-XXL encoder, and the resulting embeddings are injected via cross-attention into all layers of the Video-DiT. This allows semantic motion cues to modulate the generation process, ensuring alignment between the synthesized video and the ego-vehicle’s dynamic constraints.

\section{Additional Experimental Results}

\subsection{Qualitative analysis of latent representations.}
\begin{figure*}[t] 
    \centering \includegraphics[width=\linewidth]{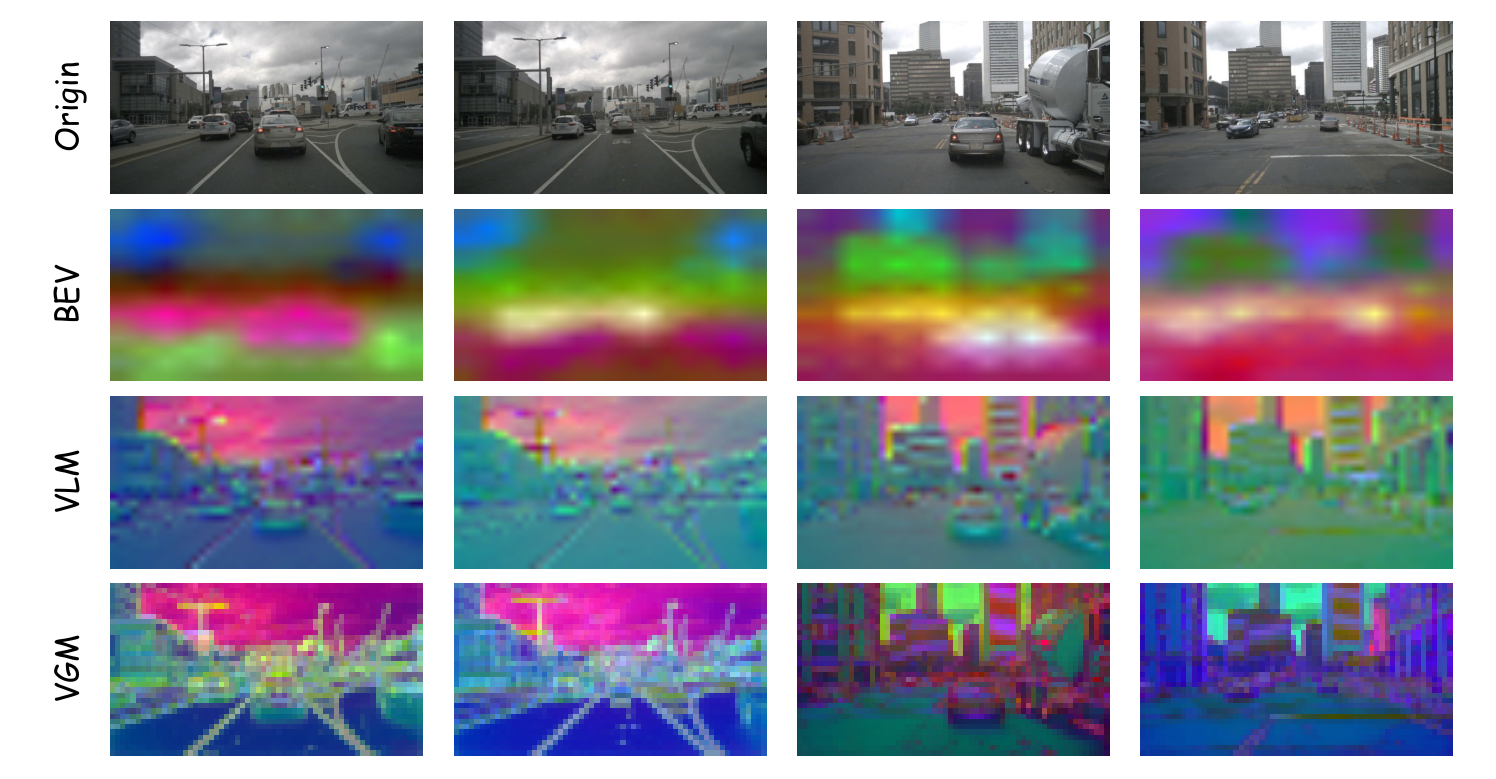} 
    \caption{\textbf{Qualitative analysis of latent representations.} We visualize the quality of latent representations from three different feature sources: perspective-view features extracted from BEVFormer~\cite{li2024bevformer}'s ResNet-101 backbone, VLM features from the pretrained Qwen2.5-VL model in ReCogDrive~\cite{li2025recogdrive}, and VGM (Video Generation Model) features from our DriveLaW-Video. To enable visual comparison, we apply PCA to reduce each representation to its top 3 principal components and map them to RGB channels. From top to bottom, each row displays: (1) the original input frame, (2) BEV features, (3) VLM features, and (4) VGM features, all upsampled to 1280×704 for visualization. While the BEV and VLM features appear diffuse, unstable, and exhibit irregular focus shifts, our VGM features are notably sharper, contain significantly less noise, and demonstrate superior semantic coherence with strong spatial structure awareness, even under severe driving motion.
}
    \label{fig:feature} 
\end{figure*}

To demonstrate that VGM (Video Generation Model) latent features can serve as more efficient and informative conditions for action learning, we conduct a systematic analysis. As shown in Fig.~\ref{fig:feature}, we visualize and compare three types of latent representations. We apply PCA (Principal Component Analysis)~\cite{abdi2010principal} to project each representation to 3 principal components mapped to RGB channels, all upsampled to 1280×704 (Note that for BEV features, limited by the single-view visual input, we extract intermediate backbone features before the BEV query transformation to ensure fair comparison). The visualization clearly shows that BEV and VLM features are diffuse, unstable, and exhibit irregular focus patterns. In contrast, VGM features are sharper, less noisy, and demonstrate superior semantic coherence with strong spatial structure awareness, even under challenging driving conditions. This suggests that VGM features provide a more suitable representation for action learning in autonomous driving.

\subsection{Quantitative Evaluation of Video Generation}
\label{sec:video_quantitative}

Tab.~\ref{tab:videogen} presents extensive quantitative evaluation of video generation quality across multiple datasets and horizon lengths.

\begin{table}[h]
\centering
\caption{\textbf{Quantitative evaluation of video generation.} We report FID and FVD on NuScenes and OpenDV, and FVD at varying horizon lengths on NuPlan.}
\setlength{\tabcolsep}{4pt}
\scalebox{0.7}{
\begin{tabular}{@{}l|cc|cc|cccc@{}}
\toprule
\textbf{Methods} & \multicolumn{2}{c|}{\textbf{NuScenes}} & \multicolumn{2}{c|}{\textbf{OpenDV}} & \multicolumn{4}{c}{\textbf{NuPlan }} \\
\cmidrule(lr){2-3} \cmidrule(lr){4-5} \cmidrule(lr){6-9}
                 & \textbf{FID}$\downarrow$ & \textbf{FVD}$\downarrow$ & \textbf{FID}$\downarrow$ & \textbf{FVD}$\downarrow$ & \textbf{FVD}$_{24}$$\downarrow$ & \textbf{FVD}$_{40}$$\downarrow$ & \textbf{FVD}$_{80}$$\downarrow$ & \textbf{FVD}$_{100}$$\downarrow$ \\
\midrule
Epona            & 7.5 & 82.8 & 6.9 & 80.7 & 61.3 & 74.9 & 239.6 & \textbf{277.3}  \\
\textbf{DriveLaW}& \textbf{4.6} & \textbf{81.3} & \textbf{4.6} & \textbf{72.9} & \textbf{55.6} & \textbf{71.2} & \textbf{230.2} & 296.1 \\

\bottomrule
\end{tabular}
}
\label{tab:videogen}
\end{table}

\noindent\textbf{Cross-Dataset Generalization.} 
In Tab.~\ref{tab:videogen}, we evaluate zero-shot generalization on the OpenDV dataset following GEM (CVPR 2025). DriveLaW-Video outperforms Epona on OpenDV, indicating robust generalization beyond the training domains of NuScenes/NuPlan.

\noindent\textbf{Long-Horizon Generation.} 
Tab.~\ref{tab:videogen} reports FVD at varying prediction horizons (24, 40, 80, and 100 frames) on NuPlan. DriveLaW consistently outperforms Epona up to 80 frames and Epona shows better performance at 100 frames. Moreover, Epona is substantially slower (Tab.~\ref{tab:runtime}), and this gap increases with horizon length. Considering both quality and efficiency, DriveLaW provides a more practical trade-off for realistic driving horizons.




\subsection{Inference Speed Analysis}
To evaluate the efficiency of our video generation stage, we compare the per-frame speed of DriveLaW with the unified world-model baseline Epona under identical experimental settings: single NVIDIA 4090 GPU, 30 DiT sampling steps, and matching resolutions as listed in Tab.~\ref{tab:speed_video}. 

\begin{table}[h]
    \centering
    \caption{Video generation speed per frame on a single NVIDIA 4090 GPU with 30 DiT sampling steps.}
    \label{tab:speed_video}
    \begin{tabular}{lccc}
        \toprule
        Method & Resolution &  Params & Times   \\
        \midrule
        Epona & $1024	\times512$ & $\sim 1.9$B        & 0.88\,s \\ 
        \midrule
        \multirow{3}{*}{\textbf{DriveLaW (Ours)}} 
            & $768	\times512$   &                     & 0.12\,s \\  
            & $1024	\times512$  & $\sim 2.0$B         & 0.18\,s \\ 
            & $1280	\times704$  &                     & 0.39\,s \\ 
        \bottomrule
    \end{tabular}
\end{table}

As shown in Tab.~\ref{tab:speed_video}, DriveLaW achieves substantially faster generation at lower resolutions. For $768\times512$, DriveLaW requires only 0.12\,s per frame, while at $1024	\times512$ the speed remains modest at 0.18\,s, despite the model size being slightly larger than Epona's. 
At the highest resolution ($1280\times704$), DriveLaW achieves 0.39\,s per frame, which is more than twice as fast as Epona’s result 0.88\,s, even though our output resolution is significantly higher.

These results indicate that the proposed architectural optimizations, which include a higher compression ratio and hybrid decoding, preserve runtime efficiency across resolutions. This allows DriveLaW to deliver competitive generation speed while maintaining high video fidelity. 

\subsection{Runtime Performance on H20 GPU}
\label{sec:runtime_h20}

For completeness, we also report inference speed on an NVIDIA H20 GPU in Table~\ref{tab:runtime}. 

\begin{table}[h]
    \centering
    \caption{\textbf{Runtime performance on an NVIDIA H20 GPU.} We report trajectory planning time, and per-frame video generation time.}
    \label{tab:runtime}
    \scalebox{0.8}{
    \begin{tabular}{lccccc}
        \toprule
        Method & Resolution & Params  & Traj. (s) & Frame (s) \\
        \midrule
        Epona & $1024\times512$ & $\sim 1.9$B  & \textbf{0.42} & 1.06 \\ 
        \textbf{DriveLaW (Ours)} & $1024\times512$ & $\sim 2.0$B  & 0.71 & \textbf{0.21} \\ 
        \bottomrule
    \end{tabular}
    }
\end{table}

\subsection{Ablation on Noise Reinjection Usage}
We conduct an ablation study to evaluate the effect of enabling or disabling the proposed noise reinjection mechanism on video generation quality.
The experiments are performed on the nuScenes validation set, with FID and FVD as evaluation metrics.

\begin{table}[h]
    \centering
    \caption{Effect of enabling noise reinjection on driving video generation quality.}
    \label{tab:reinjection_usage}
    \begin{tabular}{lcc}
        \toprule
        Setting & FID $\downarrow$ & FVD $\downarrow$ \\
        \midrule
        w/o Noise Reinjection & 6.1 & 102.1 \\
        \textbf{w/ Noise Reinjection (Ours)} & \textbf{4.6} & \textbf{81.3} \\
        \bottomrule
    \end{tabular}
\end{table}

As shown in Tab.~\ref{tab:reinjection_usage}, removing noise reinjection results in a noticeable degradation in temporal coherence and a slight decline in spatial fidelity. By selectively perturbing high-frequency regions before each denoising step, our method compels the generator to actively regenerate fine details, thereby improving both sharpness and temporal stability while reducing artifacts.

\section{More Qualitative Results}
In this section, we present additional qualitative examples to further illustrate the capabilities of DriveLaW in diverse driving scenarios and planning tasks. 

\begin{figure*}[p] 
    \centering
    \includegraphics[width=\textwidth]{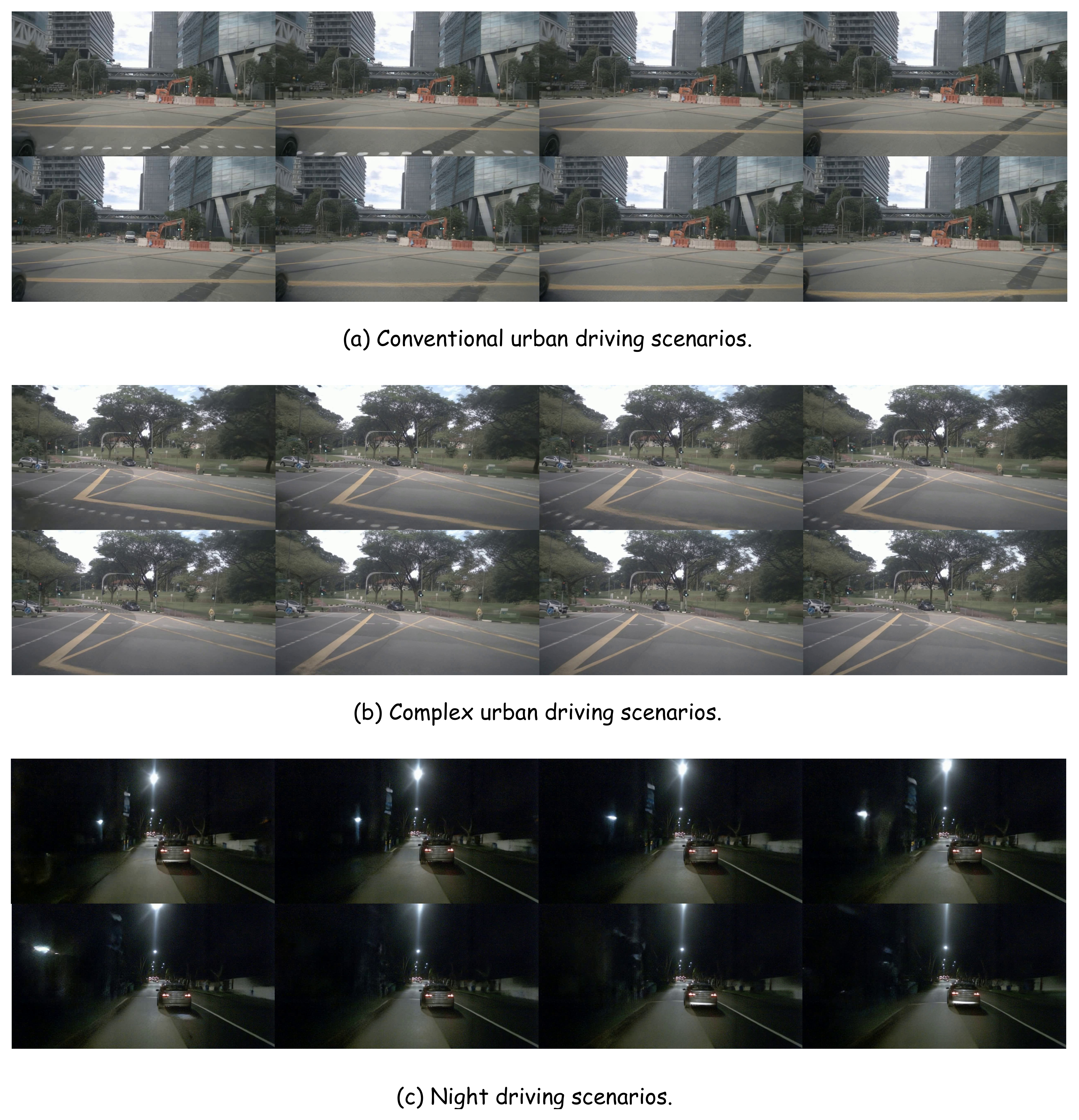}
    \caption{
    \textbf{Qualitative examples of DriveLaW video generation on the nuScenes dataset.}
    (a) Conventional urban driving scenarios, showing stable lane keeping and interactions with surrounding traffic. 
    (b) Complex urban driving scenarios involving dense multi-agent interactions, turning maneuvers, and occlusions. 
    (c) Night driving scenarios, demonstrating the model’s robustness to low-light conditions while preserving temporal consistency and fine details.
    }
    \label{fig:nuscenes_qualitative}
\end{figure*}

\subsection{Video Generation on nuScenes}

We evaluate DriveLaW on the nuScenes validation set across a wide variety of real-world driving scenarios. As shown in Fig.~\ref{fig:nuscenes_qualitative}, our results demonstrate that the model maintains temporal coherence, fine-grained spatial detail, and robust performance across diverse visual conditions.

\subsection{Planning Results Visualization}

\begin{figure*}[p] 
    \centering
    \includegraphics[width=\textwidth]{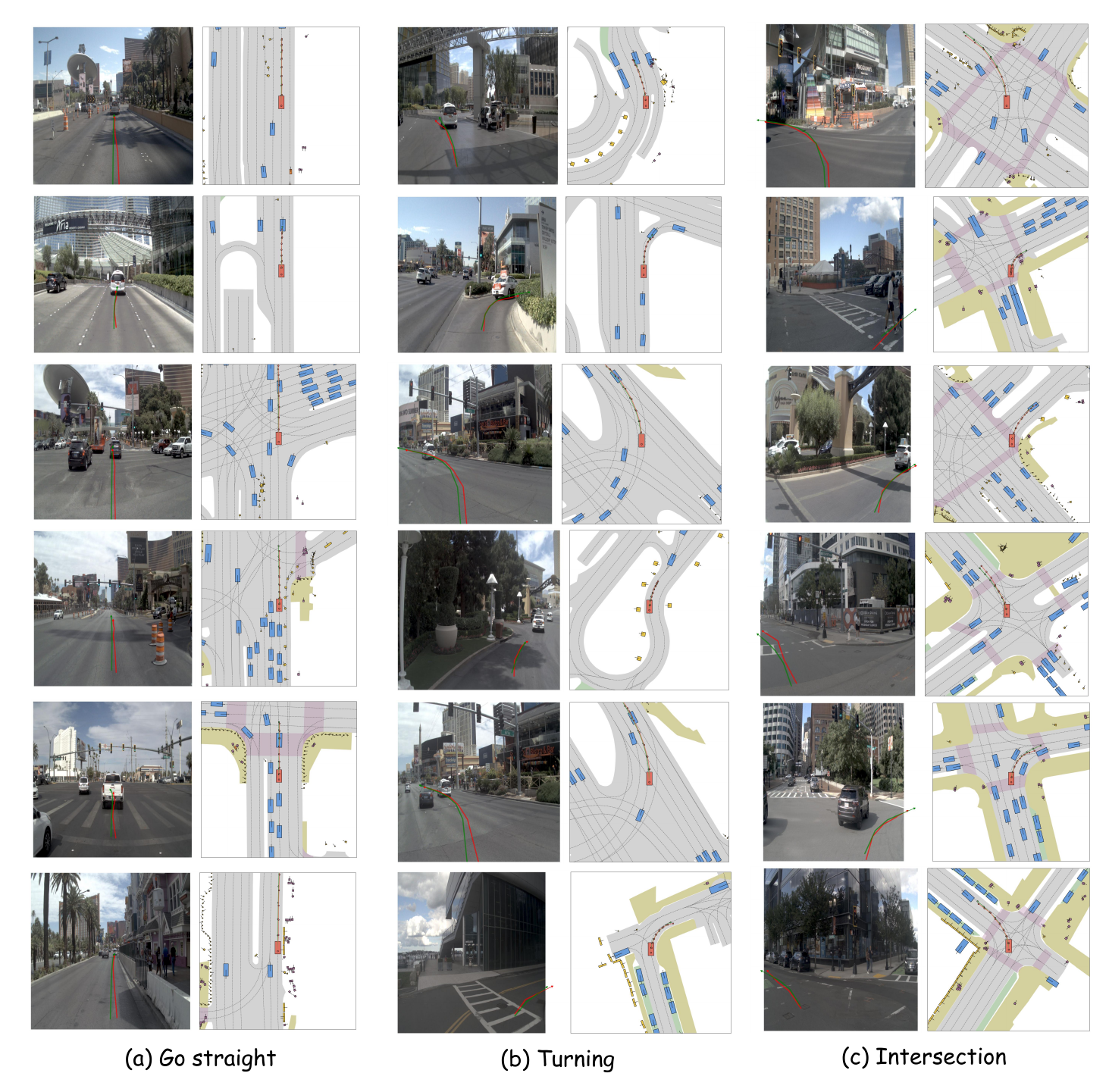}
    \caption{
    \textbf{Qualitative results on the Navtest benchmark. }}
    \label{fig:navtest_qualitative}
\end{figure*}

As shown in Fig.~\ref{fig:navtest_qualitative}, we present representative cases from the Navtest splits, highlighting DriveLaW’s ability to predict future trajectories while ensuring safety and smoothness.

\subsection{Supplementary Video Demonstrations}

To facilitate clearer understanding, we provide 6 MP4 demo videos in the supplementary material, including 4 normal-driving scenarios and 2 rainy-weather scenarios. These examples help reviewers and readers visually assess temporal consistency, spatial detail, and the practical utility of planning outputs in diverse and challenging driving conditions.

\section{Limitations and Future Work}
\label{sec:limitations}

While DriveLaW demonstrates strong performance in video generation and trajectory planning, we acknowledge several limitations that present opportunities for future research.

\subsection{Motion Artifacts in High-Compression VAE.}
To achieve efficient inference, DriveLaW employs a high-compression Video-VAE with a $32 \times 32 \times 8$ downsampling factor. Our experiments reveal that such aggressive compression introduces noticeable artifacts during reconstruction, particularly in high-motion scenarios. These artifacts propagate to the video generation stage, manifesting as visual distortions during rapid ego-motion or dynamic agent interactions. Although our proposed noise reinjection mechanism mitigates this issue to some extent (Tab.~\ref{tab:reinjection_usage}), it does not fundamentally resolve the underlying limitation. We plan to address this through architectural improvements and advanced training strategies in future work.

\subsection{Inference Latency.}
Despite our optimizations (\eg, high-compression VAE, resolution scaling, and hybrid decoding), DriveLaW's inference speed remains slower than end-to-end planning models that bypass explicit video generation. This gap stems from the inherent computational demands of diffusion-based video generation. 

\subsection{Scalability and Future Outlook.}
Notwithstanding these limitations, DriveLaW's primary advantage lies in its \emph{scalability} with rapid advances in video generation technology. As foundational video models continue to improve in quality, speed, and efficiency, DriveLaW's performance will advance commensurately without requiring architectural redesign. Furthermore, the paradigm enables the research community to leverage powerful pretrained video generators to rapidly develop generalizable planners with minimal domain-specific training. With anticipated improvements in inference acceleration techniques (\eg, distillation, quantization, and dedicated hardware), we envision the DriveLaW paradigm becoming viable for onboard deployment in the near future.